\documentclass{article}

\usepackage[final]{corl_2019} 

\usepackage[utf8]{inputenc} 
\usepackage[T1]{fontenc}    
\usepackage{hyperref}       
\usepackage{url}            
\usepackage{booktabs}       
\usepackage{amsfonts}       
\usepackage{nicefrac}       
\usepackage{microtype}      

\usepackage{graphicx}
\usepackage{amsmath}
\usepackage{amssymb}
\usepackage{graphics} 
\usepackage{graphicx} 
\usepackage{mathptmx} 
\usepackage{times} 
\usepackage{bm}
\usepackage{newtxmath}
\usepackage{newtxtext}
\usepackage{array}
\usepackage{soul}
\usepackage[]{algorithm2e}
\usepackage{xspace}
\usepackage[usenames,dvipsnames]{xcolor}
\usepackage[bottom]{footmisc}
\usepackage{pgfplots}
\pgfplotsset{compat=newest}
\usepgfplotslibrary{groupplots}
\usepgfplotslibrary{dateplot}
\usepackage{tabu}
\usepackage{array}
\usepackage{stackengine}

\usepackage{amsmath,amsfonts,bm}

\def\eqref#1{equation~\ref{#1}}
\def\Eqref#1{Equation~\ref{#1}}

\def\1{\bm{1}}

\DeclareMathAlphabet{\mathsfit}{\encodingdefault}{\sfdefault}{m}{sl}
\SetMathAlphabet{\mathsfit}{bold}{\encodingdefault}{\sfdefault}{bx}{n}

\DeclareMathOperator*{\argmin}{arg\,min}

\DeclareMathOperator{\sign}{sign}

\DeclareMathOperator{\w}{\mathbf{w}}
\DeclareMathOperator{\x}{\mathbf{x}}
\DeclareMathOperator{\act}{\mathbf{u}}

\DeclareMathOperator{\Vs}{V^{*}}
\DeclareMathOperator{\dVs}{V^{*}_{\x}}

\DeclareUnicodeCharacter{00A0}{~}
\DeclareUnicodeCharacter{221E}{$\infty$}

\DeclareMathOperator{\relu}{relu}

\title{HJB Optimal Feedback Control with Deep Differential Value Functions and Action Constraints}

\author{Michael Lutter$^{1}$, Boris Belousov$^{1}$, Kim Listmann$^{2}$, Debora Clever$^{1, 2}$, Jan Peters$^{1, 3}$\\
$^1$ Department of Computer Science, Technische Universit\"at Darmstadt, Germany\\
$^2$ ABB Corporate Research Center at Ladenburg, Germany\\
$^3$Robot Learning Group, Max Planck Institute for Intelligent Systems,T\"ubingen, Germany\\
\texttt{\{lutter, belousov, peters\}@ias.informatik.tu-darmstadt.de} \\
\texttt{\{kim.listmann, debora.clever\}@de.abb.com} \\
}

\begin{document}

\maketitle

\begin{abstract}
Learning optimal feedback control laws capable of executing optimal trajectories is essential for many robotic applications. Such policies can be learned using reinforcement learning or planned using optimal control. While reinforcement learning is sample inefficient, optimal control only plans an optimal trajectory from a specific starting configuration. 
%
In this paper we propose HJB control to learn an optimal feedback policy rather than a single trajectory using principles from optimal control. By exploiting the inherent structure of the robot dynamics and strictly convex action cost, we derive principled cost functions such that the optimal policy naturally obeys the action limits, is globally optimal and stable on the training domain given the optimal value function. The corresponding optimal value function is learned end-to-end by embedding a deep differential network in the Hamilton-Jacobi-Bellmann differential equation and minimizing the error of this equality while simultaneously decreasing the discounting from short- to far-sighted to enable the learning.
%
Our proposed approach enables us to learn an optimal feedback control law in continuous time, that in contrast to existing approaches generates an optimal trajectory from any point in state-space without the need of replanning. The resulting approach is evaluated on non-linear systems and achieves optimal feedback control, where standard optimal control methods require frequent replanning.
\end{abstract}

\section{Introduction} \vspace{-3mm}
Specifying a task through a reward function and letting an agent autonomously discover a corresponding controller
promises to simplify programming of complex robotic behaviors by reducing the required amount of manual engineering. Previous research demonstrated that such approach can successfully generate robot controllers capable of performing dexterous manipulation \cite{mordatch2012contact, andrychowicz2018learning, toussaint2018differentiable} and locomotion \cite{mordatch2012discovery, heess2017emergence}. These controllers were obtained via reinforcement learning or trajectory optimization. While reinforcement learning optimizes a possibly non-linear policy $\pi$ under the assumption of unknown rewards, dynamics and actuation limits, 
trajectory optimization plans a sequence of $n$ actions 
and states 
using the known model, reward function, initial state and actuator limits. 
When applied to the physical system, the planned trajectories must be augmented with a hand-tuned tracking controller to compensate modeling errors. 

To obtain a globally optimal feedback policy that naturally obeys the actuator limits without randomly sampling actions on the system as in reinforcement learning, we propose to incorporate actuator limits within the cost function and obtain the optimal feedback controller by embedding a deep differential network and the known model in the Hamilton-Jacobi-Bellman (HJB) differential equation. The network weights are learned using a curricular learning scheme that adapts the discounting from short to far-sighted to ensure learning of the optimal policy despite the multiple spurious solutions of the HJB. Assuming the inherent structure of most robotic tasks, i.e., control-affine dynamics due to holonomicity of mechanical systems and perfect approximation of the value function, the learned policy is globally optimal on the state domain $\Omega$, guaranteed to be stable and does not require any replanning or hand-tuning of the feedback gains. Incorporating the actuation limits within the cost function, transforms the constrained optimization problem to an unconstrained problem. Thereby enabling the learning of feasible feedback policies using gradient descent.  

Our technical contributions are the following. First, the derivation of strictly convex cost functions that make common classes of controllers, including torque limited policies, optimal. Second, the derivation of a curricular learning scheme that can learn the optimal value function and optimal feedback policy by minimizing the residual of the HJB. Furthermore, we provide intuitive explanations about the convergence of the proposed curricular learning scheme.  

In the following, we introduce the HJB differential equation and simplify the HJB using the inherent structure of robotic problems (Section \ref{sec:hjb}). Afterwards, we derive the principled cost functions that enable torque limited policies to be optimal (Section \ref{sec:convex}). Using these insights, we propose a curriculum learning scheme to learn the optimal value function and policy in Section \ref{sec:deep}. Finally, our proposed algorithm is applied to non-linear systems (Section \ref{sec:experiments}) and the differences to existing literature is highlighted (Section \ref{sec:related}).

\section{The Hamilton-Jacobi-Bellman Differential Equation} \label{sec:hjb} \vspace{-3mm}
The optimal policy $\pi^*$ solves a task by choosing the optimal actions $\act^*$ given the current state $\x$, such that the total accumulated cost is minimized. Therefore, the optimal policy $\pi^{*}$ is described by
\begin{align} \label{eq:opt_problem}
    \pi^{*}(\x, t) = \argmin_{\pi} \: \int^{T}_{t} \text{e}^{-\rho (\tau - t)} \:\: c(\x(\tau), \: \act(\tau)) \:\: \text{d}\tau \ \hspace{5pt} \text{s.t.} \hspace{5pt} \dot{\x} = f(\x, \act), \hspace{5pt}
    \x \in \Omega \subseteq \mathbb{R}^{n_x}, \hspace{5pt}
    |\act| \leq \act_{+}
\end{align}
with the action $\act = \pi(\x, t)$, the cost function $c(\x, \act)$, the dynamics $f(\x, \act)$, the discounting factor $\rho$, the state domain $\Omega$, the positive actuator limit $\act_{+}$ and the time horizon $T$. Let the optimal value function $V^{*}(\x, \: t)$ be defined as the minimum accumulated cost-to-go
, i.e.,
\begin{align} \label{eq:opt_val}
  V^{*}(\x,\: t) = \min_{\act} \: \int^{T}_{t} \text{e}^{-\rho (\tau - t)} \:\: c(\x(\tau), \act(\tau)) \:\: \text{d}\tau.
\end{align}
For the infinite time horizon, i.e., $T \rightarrow \infty$, the optimal value function is independent of time \citep{doya2000reinforcement}.
Using \Eqref{eq:opt_val}, the Hamilton-Jacobi-Bellman differential equation, the continuous counterpart of the discrete Bellmann equation, can be derived by substituting the value function at time $t' = t + \Delta t$, approximating $V^{*}(x(t'), \: t')$ with its 1st order Taylor expansion and taking the limit $\Delta t \rightarrow 0$ \citep{doya2000reinforcement}. Therefore, the HJB differential equation is described by
\begin{align}
%
%
\rho \: V^{*}(\x,\: t) &= \min_{\act} \:\: c(\x, \act) \:\: + f(\x, \act)^{T} \: \frac{\partial V^{*}}{\partial \x}. \label{eq:hjb} 
\end{align}
In the following, we will refer to $\partial\Vs/\partial\x$ as $V^{*}_x$. The HJB has multiple solutions and only incorporating the boundary condition 
\begin{align} \label{eq:bnd}
  f(\bar{\x}, \act^{*})^{T} \bm{\eta}(\bar{\x}) \leq 0 \hspace{15pt} \text{for} \hspace{5pt} \bar{\x} \in \partial \Omega
\end{align}
with the outward pointing normal vector $\bm{\eta}$ defined on the state domain boundary $\partial \Omega$ makes the solution unique \cite{fleming2006controlled}. This boundary condition implies that the optimal action always prevent the system to leave the state domain. Within the LQR-setting, this boundary condition implies the positive-definiteness of the quadratic value function. 
Using the control affine dynamics of mechanical systems with holonomic constraints, i.e., $\dot{\x} = f(\x, \: \act) = \mathbf{a}(\x) + \mathbf{B}(\x)\act$,
\Eqref{eq:hjb} and \Eqref{eq:bnd} can be further simplified to
\begin{align}
\label{eq:initial_hjb}
    \rho \: \Vs &= \mathbf{a}^{T}(\x) \dVs \: + \: \min_{\act} \left[\: c(\x, \act) +  \act^{T} \mathbf{B}^{T}(\x) \dVs  \right], \hspace{25pt} \big(\mathbf{a}(\bar{\x}) + \mathbf{B}(\bar{\x}) \act^{*} \big)^{T} \bm{\eta}(\bar{\x}) \leq 0.
\end{align}

\section{Incorporating Action Constraints within the Cost Function} \label{sec:convex} \vspace{-3mm}
Assuming a separable cost, i.e., a separate cost function for state and action, we show that the optimization problem on the right-hand side of \Eqref{eq:initial_hjb} can be solved in closed form if the action cost is strictly convex. Moreover, action constraints can be naturally accommodated if the action cost has a barrier shape. Thereby, the constrained optimization in \Eqref{eq:opt_problem} is transformed into an unconstrained problem, with the optimal policy inherently obeying the action limits.

\vspace{-3mm}
\subsection{Strictly Convex Action Cost} \vspace{-3mm}
Let the separable cost be defined as $c(\x, \: \act) = r(\x) + g(\act)$,
with the task dependent state cost $r(\x)$ and the strictly convex action cost $g(\act)$.
The strong convexity of $g$ generalizes the positive definiteness assumption on the action cost required by LQR. 
Under these assumptions, the HJB equation becomes
\begin{align}
    \rho \: \Vs &= \min_{\act} \left[ g(\act) + \act^{T} \mathbf{B}^{T}(\x) \dVs \right] + r(\x) + \mathbf{a}^{T}(\x) \dVs.
\end{align}
The optimal action and the optimal policy can be computed in closed form by employing the convex conjugate function $g^{*}(\w) = \sup_{\act} \{ \act^T \w - g(\act) \}$ and exploiting its defining property $\nabla g^{*} = (\nabla g)^{-1}$,
\begin{equation}
    \frac{\partial}{\partial \act} \left[ g(\act) + \act^{T} \mathbf{B}^{T}(\x) \dVs \right] =\nabla g(\act) + \mathbf{B}^{T}(\x) \dVs \coloneqq 0
    \quad\Rightarrow\quad \act^{*} = \nabla g^{*}(-\mathbf{B}^{T}(\x) \dVs).
    \label{eq:opt_policy}
\end{equation}
The strict convexity of $g$ assures that \Eqref{eq:opt_policy}
provides a unique global minimum~\cite{boyd2004convex}. Importantly, \Eqref{eq:opt_policy} describes the optimal policy in closed form and therefore no learning of the policy is required. The value function is also a Lyapunov function and hence, the policy is stable \citep{liberzon2011calculus}. Substituting the optimal action $\act^{*}$ into the HJB and using the Fenchel-Young identity $g(\nabla g^{*}(\w)) + g^{*}(\w) = \w^{T} \nabla g^{*}(\w)$ with $\w = -\mathbf{B}^{T}(\x) \dVs$, we arrive at the final form of the HJB equation
\begin{align}
    %
    \rho \: \Vs &= \: r(\x) +  \mathbf{a}^{T}(\x) \dVs - g^{*}\left(-\mathbf{B}^{T}(\x) \dVs \right). \label{eq:final_hjb}
\end{align}
Notably, the HJB equation in the form of \Eqref{eq:final_hjb} is a straightforward equality and does not contain a nested optimization problem in contrast to the original formulation in \Eqref{eq:initial_hjb}. Therefore, one only needs to find the optimal value function and then the optimal feedback controller is directly given.

\begin{table}[t]
  \scriptsize
  \centering
  \caption{\footnotesize
    Selected action costs.
    The choice of the action cost $g(\act)$
    determines the range of actions $\act \in \textrm{dom}\,(g)$,
    as well as the form of the optimal policy $\nabla g^{*}(\w)$
    and the type of non-linearity in the HJB equation $g^{*}(\w)$.
    Section 1 of the table contains policies with standard action domains.
    Section 2 provides formulae for shifting and scaling actions and scaling costs.
    Section 3 shows how to use the formulae.
    Section 4 gives limiting policy shapes.
}
  \begin{tabular*}{\textwidth}{l l l l l}
    \toprule
      Policy Name
        & Action Range
        & Action Cost $g(\act)$
        & Policy $\nabla g^{*}(\w)$
        & HJB Nonlinear Term $g^{*}(\w)$ \\
    \midrule
        Linear 
            & $\act \in \mathbb{R}^{n_u}$
            & $\frac{1}{2} \act^{T} \mathbf{R}\act$
            & $\mathbf{R}^{-1} \w$
            & $\frac{1}{2} \w^{T} \mathbf{R}^{-1} \w$ \\
        Logistic
            & $\mathbf{0} < \act < \1$
            & $ \act^T \log \act + (\1-\act)^{T}\log(\1-\act)$
            & $\frac{\1}{\1+e^{-\w}} \eqcolon \sigma(\w)$
            & $\1^{T} \log\left(\1+e^{\w}\right)$ \\
        Atan
            & $-\frac{\pi}{2}\1 < \act < \frac{\pi}{2}\1$
            & $-\log \cos \act$
            & $\tan^{-1}(\w)$
            & $\w^T \tan^{-1}(\w)
                -\frac{1}{2} \1^{T} \log (\1 + \w^2)$ \\
    \midrule
        Action-Scaled
            & $\act \in \alpha \,\textrm{dom}\,(g)$
            & $\alpha g(\alpha^{-1}\act)$
            & $\alpha\nabla g^{*}(\w)$
            & $\alpha g^{*}(\w)$ \\
        Cost-Scaled
            & $\act \in \textrm{dom}\,(g)$
            & $\beta g(\act)$
            & $\nabla g^{*}(\beta^{-1}\w)$
            & $\beta g^{*}(\beta^{-1}\w)$ \\
        Action-Shifted
            & $\act \in \textrm{dom}\,(g) - \gamma \1$
            & $g(\act + \gamma\1) - g(\gamma\1)$
            & $\nabla g^{*}(\w) - \gamma\1$
            & $g^{*}(\w) - \gamma \1^{T} \w$ \\
    \midrule
        Tanh
            & $-\1 < \act < \1$
            & $g_{\textrm{logistic}}(\frac{\act+\1}{2})
                - g_{\textrm{logistic}}(\frac{1}{2})$
            & $\tanh \w = 2\sigma(2\w) - \1$
            & $\1^{T} \log\cosh \w$ \\
        TanhActScaled
            & $-\alpha\1 < \act < \alpha\1$
            & $\alpha g_{\tanh}(\alpha^{-1}\act)$
            & $\alpha \tanh \w$
            & $\alpha \1^{T} \log\cosh \w$ \\
        AtanActScaled
            & $-\alpha\1 < \act < \alpha\1$
            & $-\frac{2\alpha}{\pi} \log \cos (\frac{2\alpha}{\pi} \act)$
            & $\frac{2\alpha}{\pi} \tan^{-1}(\w)$
            & $\frac{2\alpha}{\pi} g_{\textrm{atan}}^*(\w)$ \\
    \midrule
        Bang-Bang
            & $-\1 \leq \act \leq \1$
            & $\chi_{[-\1,\1]}(\act),$ $\chi$ - charact. fun. 
            & $\sign \w$
            & $\| \w \|_1$ \\
        Bang-Lin
            & $-\1 \leq \act \leq \1$
            & $\frac{1}{2}\act^T \act \; \chi_{[-\1,\1]}(\act)$
            & $-\1 + \sum_{\delta = -1}^1 \relu(\1 - \delta\w)$
            & $ \1^T L_{\1} (\w),$ $L_\delta(a)$ - Huber loss  \\
    \bottomrule
  \end{tabular*}
\label{table:convex_conjugate_functions}
\end{table}

\vspace{-3mm}
\subsection{Torque Limited Optimal Policies} \vspace{-3mm}
Exploiting the closed form solution for the optimal policy given in \Eqref{eq:opt_policy} and the convex conjugacy, one can define cost functions such that the standard controllers, including torque limited controllers, become optimal. This approach is favorable compared to the naive quadratic action cost because such cost can potentially take unbounded actions. Clipping the unbounded actions is only optimal for linear systems \cite{de2000elucidation} and increasing the action cost to ensure the action limits leads to over-conservative behavior and underuse of the control range.

The shape of the optimal policy is determined by the monotone function $\nabla g^*$. Therefore, one can define any desired monotone shape and determine the corresponding strictly convex cost by inverting $\nabla g^*$ to compute $\nabla g$ and integrating $\nabla g$ to obtain the strictly convex cost function $g(\act)$. For example, the linear policy is optimal with respect to the quadratic action cost, whereas the logistic policy is optimal with respect to the binary cross-entropy cost. The full generality of this concept based on convex conjugacy is shown in Table~\ref{table:convex_conjugate_functions}, which shows the corresponding cost functions for Linear, Logistic, Atan, Tanh and Bang-Bang controllers.
Furthermore, using the rules from convex analysis~\cite{boyd2004convex},
the effects of scaling the action limits,
shifting the action range, or scaling the action cost
can be succinctly described,
as shown by Action-Scaled, Action-Shifted,
and Cost-Scaled rows in Table~\ref{table:convex_conjugate_functions}.
This enables quick experimentation by mixing and matching costs.
For example, the action cost corresponding to the Tanh policy
is straightforwardly derived using the well-known relationship
between $\tanh(x)$ and the logistic sigmoid $\sigma(x)$
given by $\tanh(x) = 2\sigma(2x) - 1$. 
Note that a formula for general invertible affine transformations can be derived,
not only for scalar scaling and shifting. 
Classical types of hard nonlinearities~\cite{ching2010quasilinear}
can be derived as limiting cases of smooth solutions.
For example, taking the Tanh action cost $g_{\tanh}$
and scaling it with $\beta \to 0$, i.e., putting a very small cost on actions
but nevertheless preserving the action limits,
results in the Bang-Bang control shape.
Taking a different limit of the Tanh policy in which scaling is performed
simultaneously with respect to the action and cost,
the resulting shape is what we call Bang-Lin and corresponds to a function
which is linear around zero and saturates for larger input values.


\section{Learning the Value Function with Differential Networks} \label{sec:deep} \vspace{-3mm}
To obtain the optimal policy, one must solve the differential equation described in \Eqref{eq:final_hjb} to obtain the optimal value function. Learning this value function using function approximation is non-trivial because the equation contains both the value function as well as the Jacobian and has multiple solutions. The first problem can be addressed, by using a differential network, which previous research used to learn the parameters of the Euler-Lagrange differential equation \citep{lutter2018deep}, while the latter prevents the naive optimization of the HJB. Therefore, we first introduce the differential network and the naive optimization loss in Section \ref{sec:diff} and describe the curricular optimization scheme to learn the unique solution in Section \ref{sec:constrained}.

\vspace{-3mm}
\subsection{Deep Differential Network} \label{sec:diff} \vspace{-3mm}
Deep networks are fully differentiable and one can compute the partial derivative w.r.t. networks at machine precision \cite{raissi2018hidden}. Therefore, deep networks are well suited for being embedded within differential equations and learning the solution end-to-end. The deep differential network architecture initially introduced by \citep{lutter2018deep}, computes the functional value and the Jacobian w.r.t. to the network inputs within a single forward-pass by adding a additional graph within each layer to directly compute the Jacobian using the chain rule. Therefore, this architecture can be naturally used to model $V$ and $V_{\x}$ and be learned end-to-end by minimizing the error of the HJB using standard deep learning techniques. In addition, this architecture enables the fast computation of the Jacobian s.t. this network can be used for real-time control loops with up to 500Hz \citep{Lutter2019Energy}. Let $\hat{V}(\x; \:\: \psi)$ be the deep network with the network parameters $\psi$ representing the approximated value function and approximated partial derivative. This deep value function can be trained by minimizing the residual of the HJB equality (\ref{eq:hjb}) described by
\begin{align}
     \hat{V}^{*}(\x; \: \psi) = \min_{\psi} \frac{1}{N} \sum_{i}^{N} \:\: \left| \: \rho \: \hat{V}(\x_i) + g^{*}\left(-\mathbf{B}^{T}(\x_i) \hat{V}_{\x_i} \right) - \: r(\x_i) - \mathbf{a}^{T}(\x_i) \hat{V}_{\x_i} \right| \label{eq:nn_opt_problem}
\end{align}
where $\x_i$ is uniformly sampled from the state domain $\Omega$.

\vspace{-3mm}
\subsection{Constrained Optimization of the Value Function} \label{sec:constrained}\vspace{-3mm}
\begin{figure*}[t]
\centering
\includegraphics[width=\textwidth]{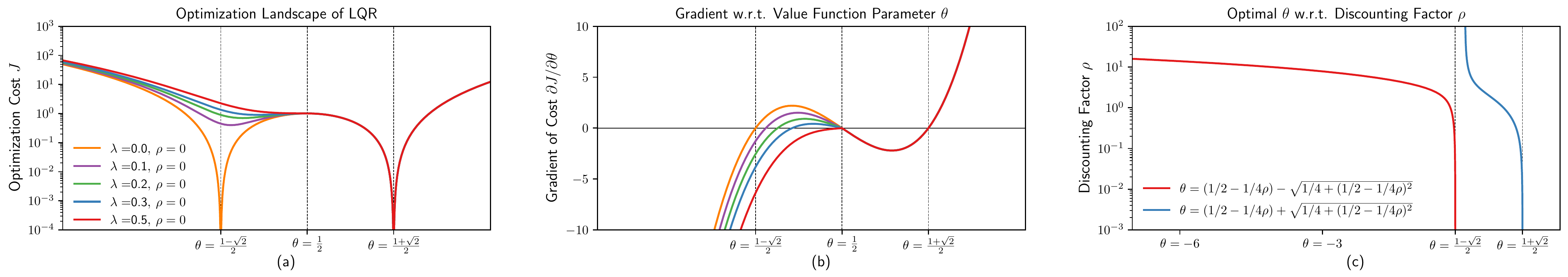}
\caption{(a) The optimization landscape for the linear system $\dot{x} = x + u$, the cost $c(x, \:u) = 1/2x^2 + 1/2u^2$ and value function $V(x) = \theta x^2$ with different penalty terms $\lambda$. For sufficiently large penalties the negative definite minima disappears. (b) The gradient $\nabla_{\theta} J$ showing the extrema for different penalty terms. (c) Optimal value function parameter $\theta$ for different discounting factors $\rho$. For $\rho \rightarrow \infty$ the negative definite parameter diverges to $-\infty$, while the positive definite solution converges to $0$.}
\label{fig:lqr_optimization}
\end{figure*}
The naive optimization of the loss of \Eqref{eq:nn_opt_problem} using gradient descent is not sufficient because the HJB has multiple solutions. Even for the one dimensional linear system with quadratic rewards and quadratic value function, i.e., $\dot{x} = Ax + Bu$,  $r(x, u) = Qx^{2} + Ru^{2}$ with $Q, R > 0$ and $V(x) = \theta x^2$, the HJB equation has two solutions described by
\begin{align*}
    V(x) &= \frac{1}{2} B^{-2} \left( (2A - \rho) R \pm \sqrt{ \left(2A - \rho \right)^2 R^2 + 4 R Q B^2} \right) x^{2}.
\end{align*}
The problem of multiple solutions can be addressed from two perspectives. First one can enforce the boundary constraint to make the solution unique. Second, one can use the a curricular learning scheme to change the discounting factor during the learning from short to far sighted. In addition, one can exploit the knowledge of terminal states and can locally clamp the value function at the terminal states to the cost function. Similar approaches have been suggested in \cite{riedmiller2005neural, tassa2007least}. 

\vspace{-3mm}
\subsubsection{Boundary Constraint} \vspace{-3mm}
The solution of the HJB for the 1d linear system is unique, when the boundary condition of \Eqref{eq:bnd} is included, i.e.,
\begin{align*}
\eta\left(\bar{x}_{+}\right) \: \left(\bar{x}_{+} - \frac{1}{2} R^{-1} V_{x}(\bar{x}_{+})\right) = 1 - \frac{1}{2} R^{-1} B^{-2} \left(\left(2A -\rho\right) R \pm  \sqrt{(2A - \rho)^2 R^2 + 4 R Q B^2}  \right) \bar{x}^{2}_{+} &\leq 0
\end{align*}
with $\bar{x}_{+} > 0$. Therefore, incorporating this constraint within the learning objective should enforce the desired optimal solution. The boundary constraint can be incorporated as additional penalty term, i.e., $\ell = \ell_{1} + \lambda \ell_b$ with
\begin{align}
    \ell_b &= 
    \max \left( 0, \:\: \big(\mathbf{a}(\bar{\x}) + \mathbf{B}(\bar{\x}) \act^{*} \big)^{T} \bm{\eta}(\bar{\x}) \right)^2
\end{align}
and optimizing this objective yields the desired optimal solution. 
For the integrator dynamics, the optimization landscape is smoothed and only one minima remains given a sufficiently large penalty term $\lambda$ (Fig. \ref{fig:lqr_optimization} a). Despite the well posed optimization surface, such constraint is hard to draft even for slightly more complex systems as the double integrator. Defining the state domain, which covers both position and velocity is non-trivial because the boundary constraint implies that the system must be controllable on the given domain boundary. Therefore, one must manually engineer the maximum region of attraction, which is dependent on the cost function $c(\x, \act)$ and their relative magnitudes. E.g., the relation of action cost to velocity cost determines the magnitude of deceleration for a moving object and hence, the slope of the boundary constraining positions and velocities. These different slopes cause the domain boundary to be specific to the cost function. Furthermore, one must account for states in which the system is not controllable. 
If wrong boundary conditions are incorporated, the optimization objectives contradict and one converges pre-maturely. Furthermore, this boundary constraint is only enforced locally and hence, the deep network may be attracted to other solutions inside the domain. 
Both drawbacks, the necessary engineering  and locality, render the boundary constraint not useful for the learning of the HJB. 

\vspace{-3mm}
\subsubsection{Changing the discounting from short to far-sighted} \vspace{-3mm}
Besides adding the boundary constraint, one can continuously decrease the discounting $\rho$ such that the value function changes from short to far-sighted solutions. This can be achieved by first initializing $\rho \gg 1$ and slowly decreasing $\rho$ once the value function is sufficiently learned. Thereby, the value function is initially attracted to a single solution and follows this solution closely through the parameter space when the discounting factor is decreased. This initial solution is unique, because taking the limit $\rho \rightarrow +\infty$ for \Eqref{eq:hjb} shows that only one unique finite value function, i.e., $V(\x) = 0, \: \forall \: \x \in \Omega$, exists. Therefore, the deep network is initially attracted to the desired optimal value functions and follows this solution closely when decreasing $\rho$. Applying this limit to the 1d linear system example, 
\begin{align*}
\lim_{\rho \rightarrow \infty} \frac{1}{2} \left( (2A - \rho) R \pm \sqrt{\left( 2A - \rho \right)^2 R^2 + 4R Q B^{2} } \right) x^{2} &= 0 \: / \: -\infty, 
\end{align*}
shows that the undesired solution diverges to $-\infty$, while the desired optimal solution approaches $\theta \rightarrow 0$. This divergence is also shown in Figure \ref{fig:lqr_optimization} c. Therefore, the value function is attracted to the desired optimal solution and follows this solution to the undiscounted infinite horizon value function. This learning scheme of changing the parameter $\rho$ during optimization can be interpreted as continuation method \citep{allgower2012numerical} and curricular learning \citep{bengio2009curriculum}. Both approaches gradually increase the task complexity to achieve faster convergence (as curriculum learning) or the avoidance of bad local optima (continuation methods). Especially, continual learning solves an initially simplified non-linear optimization problem and tracks this solutions through parameter space when the task complexity increases. Therefore, gradually decreasing $\rho$ and tracking the short-sighted $V(\x) = 0$ solution through weight space to the potentially discontinuous undiscounted infinite horizon value function is comparable to continuation methods. We only use a heuristic of decreasing $\rho$ when an error or epoch threshold is reached and aim to apply the principled methods of continuation methods for adapting $\rho$ in future work.

\begin{figure*}[t]
\centering
\includegraphics[width=\textwidth]{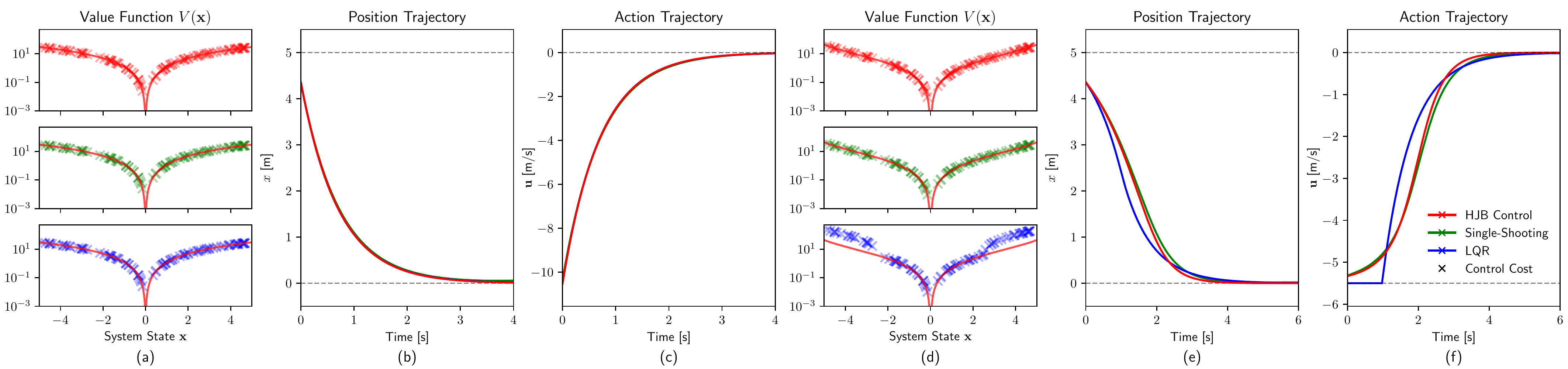}
\caption{(a) The learned value function for the linear integrator with quadratic cost. The crosses correspond to the closed-loop control cost. The achieved cost for LQR, single shooting and HJB control align with the learned value function. (b-c) State and action trajectory of the integrator with quadratic cost. LQR, single shooting and HJB control achieve optimal performance (c) The learned value function for the linear integrator with log-cosine cost incorporating the action limit. Single shooting and HJB control achieve the optimal cost, which matches the expected cost of the value function. (e-f) State and action trajectory of the integrator with log-cosine cost. Single shooting and HJB control achieve optimal control, while clipped LQR is not optimal.}
\label{fig:exp_integrator}
\vspace{-1em}
\end{figure*}

\vspace{-3mm}
\subsection{Terminal States} \vspace{-3mm}
One can exploit the knowledge of the terminal states to direct the learning of the value function. For every terminal state $\x_T$, the value function must be equal to the cost function, i.e. $\Vs(\x_{T}) = r(\x_{T})$. 
This known point of the value function can be incorporated by extending the original loss with an additional penalty term. 
In addition, this penalty term can also be added for any known regularities, e.g., constraining the the multiples of $2\pi \:k$ of a continuous revolute joint to be identical.

\section{Experiments} \label{sec:experiments} \vspace{-3mm}
The proposed approach for learning optimal feedback control is evaluated
on the stabilization of a one-dimensional linear system, the swing-up of a torque-limited pendulum and the balancing of a flexible Cartpole\footnote{The exact system descriptions including the system parameters is explained in the appendix. }. For the quadratic action cost, the actions are not limited, while the log-cosine action cost implicitly limits the actions. In the following, we will refer to our proposed approach of learning the optimal value function and applying the closed-form policy as \emph{HJB control}. The performance is compared to LQR and shooting methods\footnote{Single and multiple shooting is implemented using CasADi \citep{Andersson2018}} augmented with hand-tuned tracking controllers.

\vspace{-3mm}
\subsection{Linear System} \vspace{-3mm}
The learned value function of the linear system and the corresponding control performance is shown in Figure \ref{fig:exp_integrator}. For the quadratic cost function, HJB control, LQR and single shooting obtain identical state and action trajectories (Fig. \ref{fig:exp_integrator} b-c). For randomly sampled starting configurations, the learned value function and the accumulated cost of the HJB controller match the cost of LQR and single shooting (Fig. \ref{fig:exp_integrator} a). For the log-cosine cost, only HJB control and single shooting achieve optimal performance on the complete state domain while LQR only achieves comparable cost for limited starting configurations, when the action limits are not active. This can be seen in Figure \ref{fig:exp_integrator} d, where the expected control cost of the learned value function and the accumulated cost of the HJB controller match the cost of single shooting. In contrast, LQR achieves comparable cost for $x_0 \leq 2.5$ but significantly larger cost for $x_0 > 2.5$. 

\vspace{-3mm}
\subsection{Torque Limited Pendulum}\vspace{-3mm}
\begin{figure*}[t]
\centering
\includegraphics[width=\textwidth]{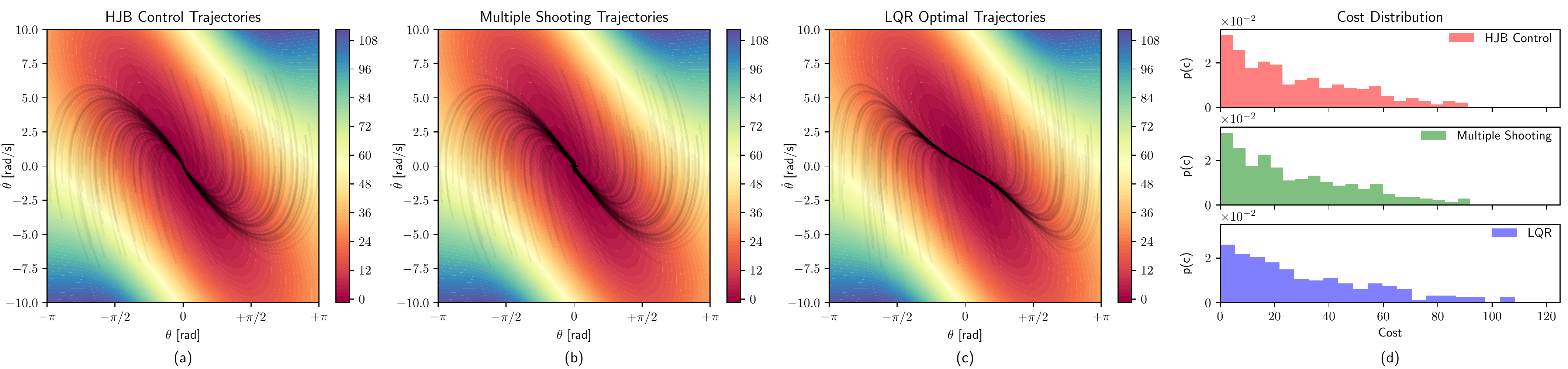}
\includegraphics[width=\textwidth]{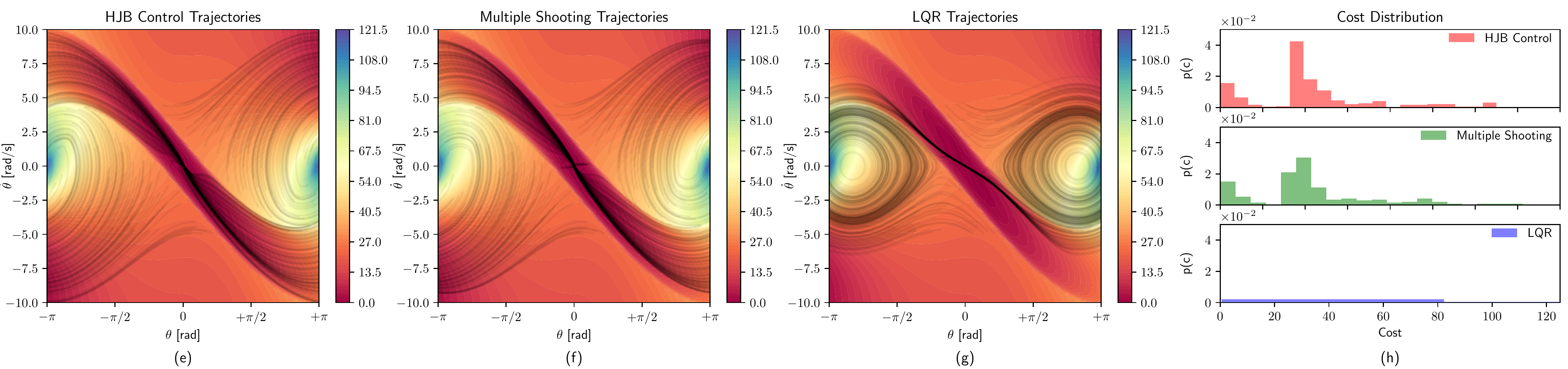}
\caption{(a-c) Learned value function for the pendulum with quadratic cost and the trajectories for HJB control (a), multiple shooting (b) and LQR (c) from 300 randomly sampled starting configurations. 
(d) Cost distributions $p(c)$ for the sampled starting configurations. The distributions for HJB control and multiple shooting are similar while the distribution of LQR requires higher cost for some starting configurations. (e-g) Learned discontinuous value function for the pendulum with log-cosine cost and the trajectories for HJB control (e), multiple shooting (f) and LQR (g). HJB control and multiple shooting can swing-up the pendulum from every sampled starting configuration, while LQR can only balance the starting configurations for which the action constraint is not active. (h) The cost distributions $p(c)$ for HJB control and multiple shooting match closely while LQR only achieves low reward for very few starting configurations.}
\label{fig:exp_pendulum}
\vspace{-1em}
\end{figure*}

The learned value function of the non-linear torque limited pendulum and the corresponding control performance of the HJB controller for both the quadratic and log-cosine cost is shown in Figure \ref{fig:exp_pendulum}. For the quadratic action cost, the value function is continuous and only locally quadratic within the surrounding of the balancing point. The trajectories of HJB control, multiple shooting and LQR from 300 randomly sampled starting configurations $\x_{0}$ are shown in Figure \ref{fig:exp_pendulum} a-c. The trajectories from HJB control match the trajectories of multiple shooting and hence are optimal for the non-linear system. In contrast, the LQR controller applies unnecessarily large actions because the system linearization at $\theta = 0$ over-estimates the system dynamics. Furthermore, the corresponding cost distribution for the starting configurations is very similar for the HJB controller and multiple shooting, while the LQR cost distributions shows that LQR requires larger cost for some starting configurations (Fig. \ref{fig:exp_pendulum} d). For the log-cosine cost, which implicitly limits the feasible actions, the value function is discontinuous and contains two ridges leading to the balancing point, because the balancing point cannot be reached directly from every point within the state domain. The learned value function and the trajectories of HJB control, multiple shooting and LQR are shown in Figure\ref{fig:exp_pendulum} e-g. Multiple shooting and HJB control achieve the swing-up from every starting configuration while LQR cannot swing-up the pendulum and only achieves balancing of starting configurations close to the upright position. Furthermore, the cost distributions for HJB control and multiple shooting match closely (Fig. \ref{fig:exp_pendulum} h). This close similarity between cost distribution shows that our proposed approach learned the optimal value function, and the corresponding feedback policy achieves optimal feedback control. 

 \vspace{-3mm}
\subsection{Flexible Cartpole}
\begin{figure*}[t]
\centering
\includegraphics[width=\textwidth]{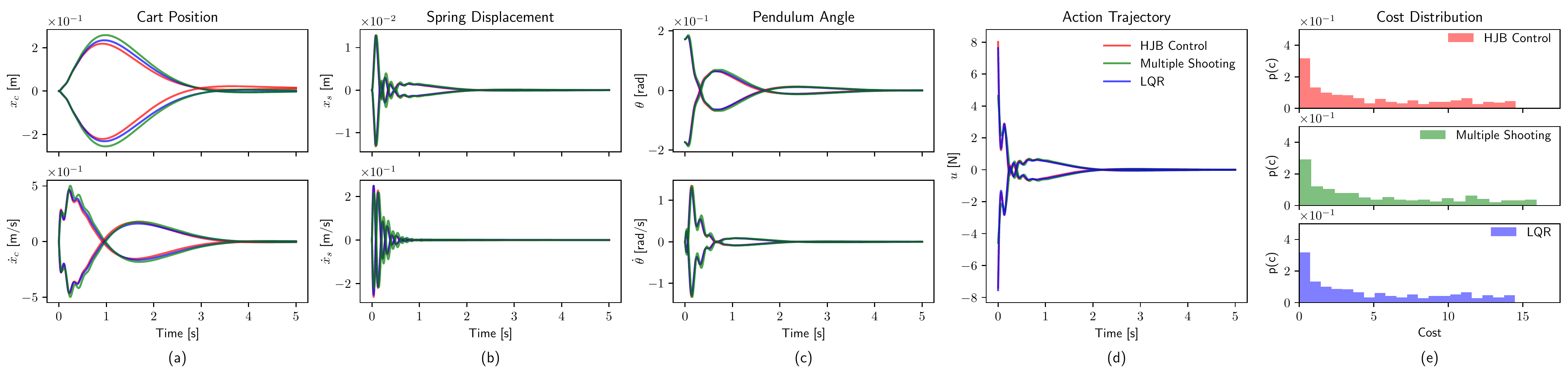}
\caption{(a-d) State and action trajectories of the flexible Cartpole from two different starting configurations using the control laws derived by HJB control, multiple shooting and LQR.
(d) Cost distributions $p(c)$ from 300 uniformly sampled starting configurations. The achieved cost for HJB control and LQR is similar while the achieved cost for multiple shooting is slightly higher due to the hand-tuned feedback gains.}
\label{fig:exp_cartpole}
\vspace{-1em}
\end{figure*}

\vspace{-3mm}
The flexible Cartpole is a Cartpole with an additional spring within the linear actuator. Therefore, the cart with the pendulum is not directly actuated and the force must be transmitted through the spring to balance the pendulum. Figure \ref{fig:exp_cartpole} shows the optimal actions computed using HJB control, LQR and multiple shooting as well as the corresponding six dimensional state trajectories consisting of cart position $x_c$, spring displacement $x_s$ and pendulum angle $\theta$. The action and state trajectories (Fig. \ref{fig:exp_cartpole} a-d) as well as the cost distribution from 300 different starting configurations (Fig. \ref{fig:exp_cartpole} e) overlap closely for HJB control and LQR, which is optimal for this linearized system and quadratic cost. Only the multiple shooting baseline does a slightly larger compensatory movement of the cart due to the hand-tuned feedback gains. These hand-tuned gains also cause the larger cost for the multiple shooting baseline visible in Figure \ref{fig:exp_cartpole} e. This similarity of LQR and HJB control shows that HJB control has learned the optimal value function and demonstrates that HJB control is also applicable to six dimensional problems.

\section{Related Work} \label{sec:related} \vspace{-3mm}
Non-linear control problems are normally solved by trajectory optimization with explicit inequality constraints on the actions \citep{bryson1975applied}. These approaches yield a single optimal trajectory, that needs to be replanned for every initial configuration and augmented with a non-optimal tracking controller. Locally-optimal tracking controller can be obtained when using iterative linear quadratic programming (iLQR) with action constraints \citep{tassa2014control} or guided policy search (GPS) \cite{levine2013guided}. In contrast to these approaches, our proposed method transforms the constrained problem to an unconstrained problem by using principled cost function and provides the globally optimal feedback controller on the domain $\Omega$. 

The transformation of the inequality constraints to the principled cost function has been explored by a number of authors. Historically the first mention of generalized convex action costs
goes back to~\cite{lyshevski1998optimal}. 
Thereafter, the $\tanh$ non-linearity derived from the corresponding action cost has been commonly used in the adaptive dynamic programming literature~\cite{abu2005nearly, yang2014reinforcement, modares2014online}. Furthermore, \cite{doya2000reinforcement} and~\cite{tassa2007least} present similar derivation to ours, with a generic convex action cost. However, these papers do not arrive at the general convex conjugate theory as these papers only treat the $\tanh$ case and do not derive the HJB in the form of \Eqref{eq:final_hjb} using the convex conjugate $g^*$.

Global optimal feedback controllers were previously learned by using the least squares solution of the HJB differential equation \cite{doya2000reinforcement, tassa2007least, yang2014reinforcement, liu2014neural}. Early on, \cite{doya2000reinforcement} used radial-basis-function networks to learn the HJB. Similarly, \citep{yang2014reinforcement, liu2014neural} learned polynomial value function using gradient descent. 
Most similar to our approach, \cite{tassa2007least} used neural networks of the Pineda architecture to learn the average cost solution of the HJB. To achieve convergence the authors added domain constraints, adapted the discounting horizon and added stochasticity to the dynamics to smoothen the value function. In contrast to the previous works, our proposed approach learns the value function using a generic deep differential network capable of representing discontinuous value functions and only requires the adaptation of the discounting factor to achieve robust convergence. Furthermore, we provide intuitive insights that show how changing the discounting factor enables the learning of the optimal value function. 
 
\section{Conclusion} \label{sec:conclusion} \vspace{-3mm}
In this paper, we showed that constrained optimization problems for finding action-limited optimal policies can be transformed into an unconstrained problem by designing principled strictly convex cost-functions, which guarantee optimal policies that naturally obey the actuator limits. Exploiting this transformation as well as the affine control dynamics of mechanical systems with holonomic constraints, we showed that the optimal policy can be described in closed form given the differentiable value function. The differentiable value function can be learned by embedding a deep differential network within the HJB and using a curricular learning approach. This curriculum changes the discounting from short- to far-sighted to achieve robust convergence to the desired value function and avoids the undesired solutions of the HJB, which violate the boundary conditions. The experiments demonstrated that our approach can learn the optimal feedback controller for non-linear and torque limited systems. Furthermore, HJB control achieves similar performance as shooting methods but does not require the replanning for different starting configurations and the hand-tuning of the gains.  

In future work, we plan to extend this work to model-based reinforcement learning by combining the HJB optimal feedback control with model learning in the form of Deep Lagrangian Networks \cite{lutter2018deep}, which ensures learning control-affine dynamics. Iterating between optimal control and model learning, the optimal policy and model are simultaneously learned. Besides the model learning aspect, we plan to develop an iterative sampling scheme to construct a domain limited to the surroundings of the current policy. This reduced domain should enable the scaling of our proposed approach to high dimensional problems where obtaining the global solution is not feasible.

\section*{Acknowledgement}
This project has received funding from the European Union’s Horizon 2020 research and innovation program under grant agreement No \#640554 (SKILLS4ROBOTS). Furthermore, this research was also supported by grants from ABB AG, NVIDIA and the NVIDIA DGX Station.

\bibliography{references.bib}

\section*{Appendix} \label{sec:appendix}
\subsection*{Analytic Derivation of the 1d LQR System} \vspace{-3mm}
Let the system dynamics, cost function and value function be defined as $\dot{x} = A x + B u$, $c(x, \: u) = Q x^{2} + R u^{2}$, $\Vs = \theta x^{2}$ and $\dVs = 2 \theta x$. Then \Eqref{eq:final_hjb} is described by
\begin{align*}
    \rho \: \Vs + g^{*}\left(-B(x) \dVs \right) - r(x) - a(x) \dVs &= 0 \hspace{15pt} \forall x \in \mathbb{X} \subseteq \mathbb{R}  \\
    \rho \: \Vs + \frac{1}{4} R^{-1} \left(B \dVs\right)^{2} - Q x^{2} - A \dVs x &= 0 \\
    \rho \theta x^{2} + \frac{1}{4} R^{-1} \left(2 B \theta x \right)^{2} - Q x^2 - 2 A \theta x^{2} &= 0 \\
    \left(R^{-1} B^{2} \theta^{2} + \left(\rho - 2 A \right) \theta - Q \right) x^{2} &= 0.
\end{align*}
Solving the quadratic equation yields the optimal parameter of the value function described by
\begin{align*}
    \theta = \frac{1}{2} B^{-2} \left(\left(2A - \rho \right) R \pm \sqrt{\left( 2A - \rho  \right)^2 R^{2} + 4 R Q B^{2}} \right).
\end{align*}

\subsection*{System Description}\vspace{-3mm}
This section describes the exact system specification used for simulation. All simulations were conducted with $500$Hz. 

\smallskip
\textbf{1d-Linear System:}
The dynamics of the linear system are described by $\dot{\x} = \mathbf{A}\x + \mathbf{B}\act$ and the state cost is quadratic $r(\x) = \x^{T} \mathbf{Q} \x$. For the experiments, a simple one-dimensional integrator with the parameters $A = B = 1$, $Q = 1/2$ and the domain $-5 \leq x \leq +5$ is used. For the quadratic cost function the actions are unconstrained, while for the log-cosine cost the actions are constrained to $\act_{+} = 5.5$. 

\smallskip
\textbf{Torque Limited Pendulum:}
The torque limited pendulum is a one-degree-of-freedom non-linear system with the joint position $\theta$, velocity $\dot{\theta}$ and torque $\act$. For the experiments the canonical system representation with the state $\x = \left[\theta, \dot{\theta} \right]$ with $\theta = 0$ being the upward pointing pendulum is used. The corresponding equations of motions are described by
\begin{align}
 \ddot{\theta} = \frac{3}{ml^2} \left[ \act - \frac{m g l}{2} \: \sin(\theta) \right], \hspace{25pt}
 \mathbf{a} = \left[\dot{\theta}, \:\: \pm \frac{3g}{2 l} \sin(\theta)\right], \hspace{25pt} 
 \mathbf{B} = \left[0, \:\: \frac{3}{m l^2}\right],
\end{align}
with the pendulum mass $m = 1$kg, the length $l = 1$m and the gravitational constant $g = -9.81$m/s$^2$. The state cost is given by $r(\x) = q_{0} \pi^2 \sin(\theta /2) + q_{1} \dot{\theta}^2$. For the log-cosine cost, the torque is constrained to $\act_{+} = 2.5$N/m. 

\smallskip
\textbf{Flexible Cartpole:}
The flexible Cartpole has in addition to the normal Cartpole a spring within the linear actuator. This flexible actuator is modelled by two separate carts connected by a spring, whereas only one cart is actuated. Therefore, the controller must excite/compensate the spring to balance the pendulum. The six dimensional system state is described by $\x = \left[x_c,\: x_s, \:\theta, \:\dot{x}_c, \:\dot{x}_s,\: \dot{\theta} \right]$, with the actuated cart position $x_c$, the spring displacement $x_s$ and the pendulum angle $\theta$. For the balancing experiments, the system is approximated using a linear system described by $\dot{\x} = \mathbf{A} \x + \mathbf{B} \act$. The system matrices are described by
\begin{align*}
   \mathbf{A} &= \left[ \begin{array}{cccccc}
   0 & 0 & 0 & 1 & 0 & 0 \\[0.75ex]
   0 & 0 & 0 & 0 & 1 & 0\\[0.75ex]
   0 & 0 & 0 & 0 & 0 & 1\\[0.75ex]
   0 & \frac{K_s}{m_a} & 0 & -\frac{B_a}{m_a} & 0 & 0\\[0.75ex]
   0 & -\frac{K_s}{m_a} - \frac{K_s}{m_a} & \frac{m_{\theta} g}{ m_p }& \frac{B_a}{m_a} - \frac{B_p}{ m_p} & -\frac{B_p}{m_p} & -\frac{B_p}{m_p l_{\theta}}\\[0.75ex]
   0 & -\frac{K_s}{m_p l_{\theta}} & \frac{g (m_{p} + m_{\theta})}{m_p l_{\theta}} & -\frac{B_p}{m_p l_{\theta}} & -\frac{B_p}{m_p l_{\theta}} & -\frac{B_{\theta} (m_{p} + m_{\theta}) }{m_{p} m_{\theta} l_{\theta}^2 } \\ \end{array}\right]  \\
   \mathbf{B}^{T} &= \left[ \begin{array}{cccccc}
   0 & 0 & 0 & \frac{1}{m_c} & -\frac{1}{m_c} & 0
   \end{array}\right],
\end{align*}
with the masses $m_a = 0.57$kg, $m_p = 0.375$kg, $m_{\theta} = 0.127$kg, the spring constant $K_s = 200$N/m, pendulum length $l_{\theta} = 0.1778$m, the viscous friction coefficients $B_a = B_p = 0.5$Ns/m, $B_{\theta} = 0.0024$Nms/rad and the gravitational constant $g = 9.81$m/s$^2$.

\end{document}